\documentclass{article}

\usepackage[preprint,nonatbib]{neurips_2023}

\usepackage[utf8]{inputenc} 
\usepackage[T1]{fontenc}    
\usepackage{hyperref}       
\usepackage{url}            
\usepackage{booktabs}       
\usepackage{amsfonts}       
\usepackage{nicefrac}       
\usepackage{microtype}      
\usepackage{xcolor}         
\usepackage{amsmath}
\usepackage{float,graphicx}
\usepackage{amssymb}

\usepackage{algorithm}
\usepackage{algpseudocode}

\title{Control-aware echo state networks (Ca-ESN) for the suppression of extreme events}

\author{%
  Alberto Racca 
    \\
  I-X and Department of Aeronautics\\
  Imperial College London\\
  \texttt{a.racca@imperial.ac.uk} \\
  \And
  Luca Magri \\
  Department of Aeronautics \\
  Imperial College London,\\
  The Alan Turing Institute\\
  \texttt{l.magri@imperial.ac.uk} \\
}

\begin{document}

\maketitle

\begin{abstract}
Extreme event are sudden large-amplitude changes in the state or observables of chaotic nonlinear systems, which characterize many scientific phenomena. 
Because of their violent nature, extreme events typically have adverse consequences, which call for methods to prevent the events from happening.
In this work, we introduce the control-aware echo state network (Ca-ESN) to seamlessly combine ESNs and control strategies, such as proportional-integral-derivative and model predictive control, to suppress extreme events. The methodology is showcased on a chaotic-turbulent flow, in which we reduce the occurrence of extreme events with respect to traditional methods by two orders of magnitude. This works opens up new possibilities for the efficient control of nonlinear systems with neural networks. 
\end{abstract}

\begin{figure}[H]
    \centering
    \includegraphics[width=.56\textwidth]{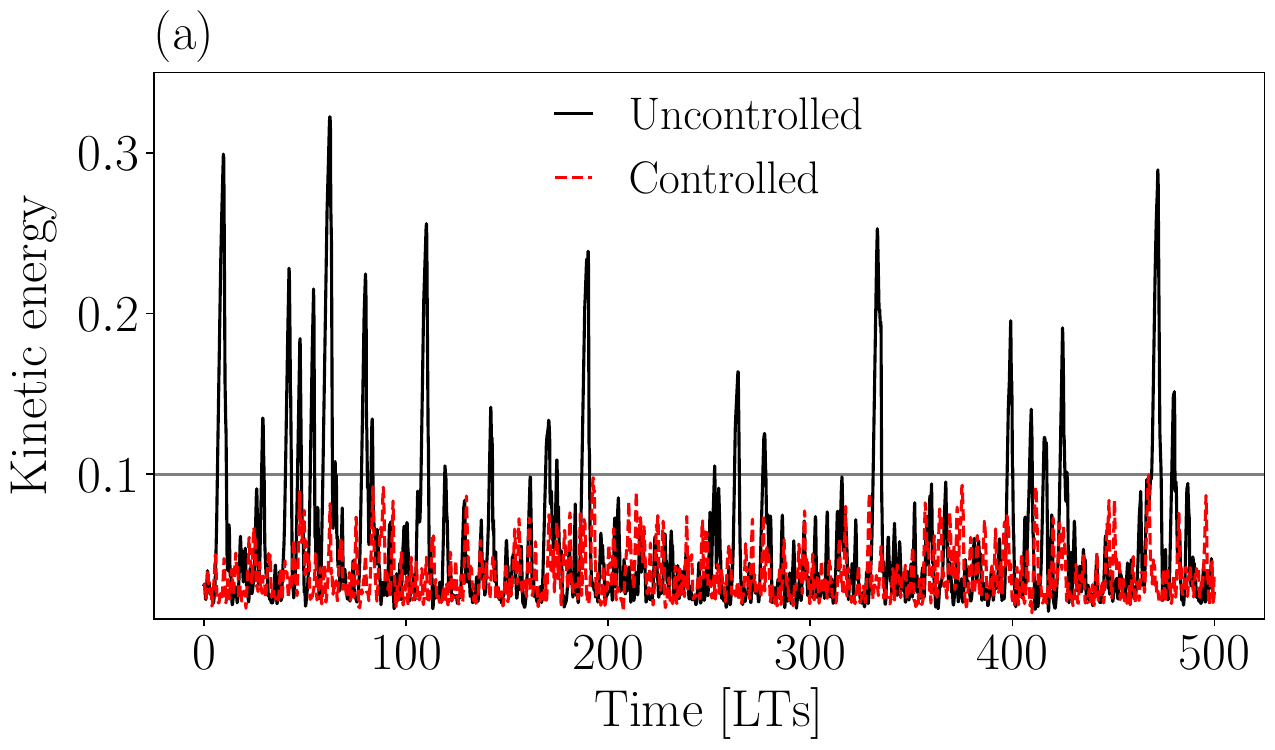}
    \includegraphics[width=.43\textwidth]{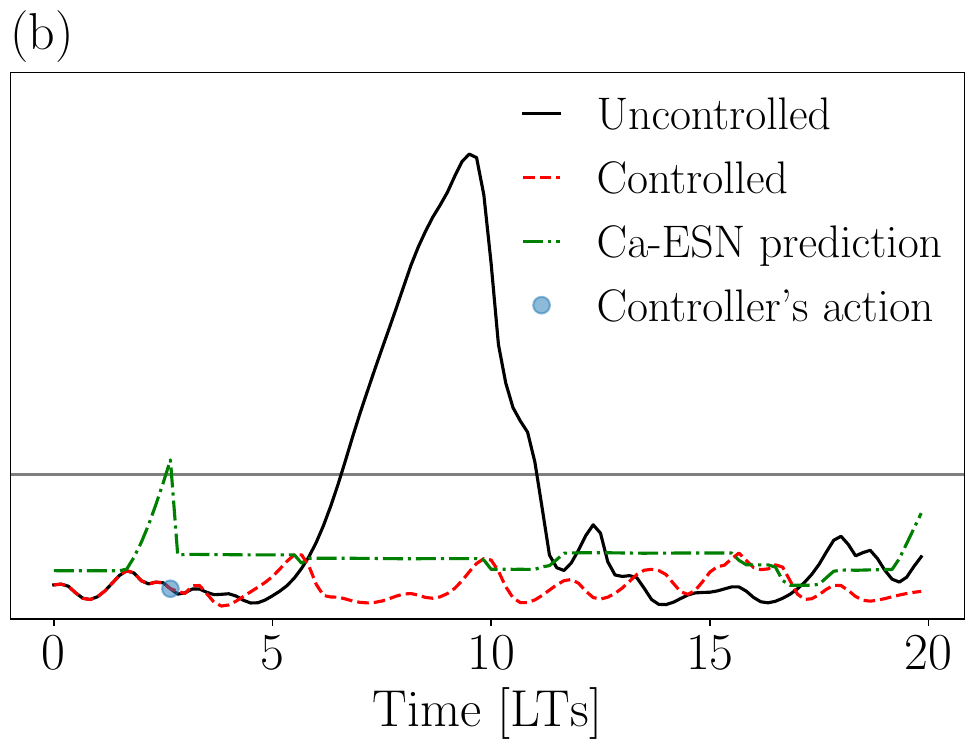}
    \caption{(a) Controlled and uncontrolled system's evolution. (b) Example of a suppressed extreme event. The controller acts before the event occurs based on the echo state network's prediction.}
    \label{fig:extreme_dyn}
\end{figure}

\section{Introduction}

Extreme events arise in a variety of natural and engineering systems, in the form of rogue waves, atmospheric events and power grid shocks, to name a few \cite{farazmand2019extreme}. These events often have negative consequences, thus, developing methods for their prediction and control is an active field of research \cite{farazmand2019extreme,sapsis2021statistics}. At the same time, the events usually show no apparent early sign of their occurrence, which makes their control from direct observations especially difficult. 
Because of this, control of extreme events is typically performed by assuming knowledge of the governing equations of the system, whose prediction via time integration activates the controller in advance \cite{nagy2007control}. 
When the governing equations are not known, data-driven modelling becomes necessary. To this end, echo state networks (ESNs) \cite{jaeger2004harnessing}, which are  state-of-the-art machines for the prediction of nonlinear dynamical systems, 
offer a promising alternative \cite{vlachas2020backpropagation}. 
ESNs have been shown to predict extreme events, and deployed to activate basic controllers for their suppression \cite{pyragas2020using,racca2022data}. 
These works, however, considered only simplistic prescribed control strategies, which have limited the applicability and performance of the controllers.
We propose the control-aware ESN (Ca-ESN) to seamlessly integrate the prediction of ESNs in the formalism of established controllers, such as proportional integral derivative (PID) controllers \cite{astrom1995pid} and model predictive control (MPC) \cite{camacho2013model, draeger1995model}, to efficiently suppress extreme events. We showcase the Ca-ESN in a model of turbulence \cite{moehlis2004low}, in which extreme events occur as intermittent burst in the total kinetic energy of the flow.

\section{The control problem: suppression of extreme events}

We analyse dynamical systems that evolve according to a set of discretized governing equations
\begin{equation}
    \mathbf{q}(t_{i+1}) = \mathbf{f}(\mathbf{q}(t_i), \mathbf{u}(t_i)),
    \label{eq:problem}
\end{equation}
where $\mathbf{q}(t_i)$ is the state of the system, which
shows extreme events in the observable, $k(t_i) = \mathbf{g}(\mathbf{q}(t_i))$ (Fig. \ref{fig:extreme_dyn}). We assume that $\mathbf{f}$ is not known, and that data on $\mathbf{q}$ is available. We wish to suppress the events through the control input, $\mathbf{u}(t_i)$, whose objective is to decrease the number of extreme events while acting as infrequently as possible on the system. 
This is translated into a quantitative goal through the average reward 
\begin{equation}
    R = \frac{1}{N}\sum_{i=1}^N r(t_i),
\end{equation}
where $r(t_i)$ is the user-defined reward at each time step. 
First, to prevent extreme events from happening, we set a negative reward, $r(t_{i,e})=-1$, for each time step, $t_{i,e}$, at which the system is experiencing an event. 
Second, to find control strategies that act rarely on the system, we select a system-dependent (smaller) negative reward, $r(t_{i,c})$, for every time step, $t_{i,c}$, in which the control strategy is activated. This is to discourage the activation of the control strategy when it is not needed. The reward is set to zero for all other time steps. 
To further characterize the controlled system, we compute the total number of time steps that the system experiences an extreme event, $N_e$, or control, $N_c$. By doing so, we analyse how often (i) the system shows extreme behaviour through the extreme events ratio, $P_{e} = N_e / N$, and (ii) control is active through the control ratio, $P_c = N_c /N$.



\section{Control-aware echo state network}

To suppress extreme events, we employ standard controllers. We first consider the proportional-integral-derivative (PID) controller \cite{astrom1995pid}, $c(k(t))$, 
\begin{equation}
        c(k(t)) = K_p k(t) + K_d \frac{d k(t)}{dt} + K_i \int_{t-\tau_i}^{t}k(t')dt';
    \label{eq:PD_control}
\end{equation}
where the proportional, $K_p$, derivative, $K_d$ and integral, $K_i$, multipliers and the integral time, $\tau_i$, are selected through Bayesian optimisation \cite{snoek2012practical} to optimize the average reward. 
Secondly, we analyse nonlinear model predictive control (MPC), which finds the optimal control sequence within a future time window \cite{camacho2013model}. 
In MPC, at every control step, $t_0$, we solve a constrained optimization problem to maximize the average reward over the future (receding) time horizon, $\tau_\mathrm{hor} = N_\mathrm{hor}dt$,
\begin{align}
    \mathop{\mathrm{max}}_{\mathbf{u}_\mathrm{opt}(t_i)} \;\; & R_\mathrm{hor} = \frac{1}{N_\mathrm{hor}}\sum_{i=1}^{N_\mathrm{hor}} r(t_i) \nonumber \\[5pt] 
    \mathrm{subject} \;\; \mathrm{to} \;\;  
    & \mathbf{q}(t_{i+1}) = \mathbf{f}(\mathbf{q}(t_i), \mathbf{u}_\mathrm{opt}(t_i))
     \;\; \mathrm{for} \;\; i \leq N_\mathrm{opt} \nonumber \\[5pt]
    &\mathbf{q}(t_{i+1}) = \mathbf{f}(\mathbf{q}(t_i), \mathbf{u}_\mathrm{fix}(t_i)), \;\;\; \mathrm{for} \;\; N_\mathrm{opt} < i < N_\mathrm{hor}
    \label{eq:mpc}
\end{align}
where  $N_\mathrm{opt} \leq N_\mathrm{hor}$ are the time steps at which we optimise the control law, $[\mathbf{u}_\mathrm{opt}(t_0),...,\mathbf{u}_\mathrm{opt}(t_0+N_\mathrm{opt}dt)]$ \cite{murray2009optimization}. The system is controlled for the remaining time steps within the time horizon by a prescribed (fixed) control law, $\mathbf{u}_\mathrm{fix} = \mathbf{0}$ (no control). 
We use two different strategies, $\mathbf{u}_\mathrm{fix}$ and $\mathbf{u}_\mathrm{opt}$, to decrease the search space and therefore the computational cost of solving \eqref{eq:mpc}. 


To enable the controllers, we propose the Control-aware Echo State Network (Ca-ESN). Echo state networks \cite{jaeger2004harnessing} nonlinearly expand the inputs into a high-dimensional reservoir, $\mathbf{r}(t_i)$, from which the output of the network is computed as a linear combination. 
The Ca-ESN provides a data-driven model for the iterative evolution of the controlled system, $\mathbf{\hat{q}}(t_{i+1}) = \mathbf{f}_\mathrm{ESN}(\mathbf{q}(t_i), \mathbf{u}(t_i))$, 
\begin{gather}
\label{eq:ESN_step}
        \textbf{r}(t_{i+1}) = \mathrm{tanh}\left(\sigma_{in}\mathbf{W}_{\mathrm{in}}\mathbf{q}_{\mathrm{in}}(t_i)+\rho\mathbf{W}\textbf{r}(t_i) + \sigma_\mathrm{c}\mathbf{W}_\mathrm{c}\mathbf{u}(t_i)\right), \nonumber \\
        \mathbf{\hat{q}}(t_{i+1}) = \mathbf{r}(t_{i+1})^T\mathbf{W}_\mathrm{out};
\label{output_mult}
\end{gather}
where the matrices $\mathbf{W}_\mathrm{in}$, $\mathbf{W}$, and $\mathbf{W}_\mathrm{c}$ are randomly generated and fixed \cite{lukovsevivcius2012practical}.  The hyperparameters, $\sigma_{\mathrm{in}}$ and $\sigma_c$ are optimized through Bayesian optimisation and recycle validation \cite{racca2021robust}. 
Because the evolution of the system is Markovian \eqref{eq:problem}, we set $\rho=0$ to eliminate the recurrence in time of the network. In this way, we simplify the architecture, thereby (i) reducing its computational cost and (ii) making it equivalent to a one-layer extreme learning machine \cite{huang2011extreme}. 
The weights of the output matrix, $\mathbf{W}_\mathrm{out}$, are the only trainable parameters. Thanks to this architecture, training the network needs only solving a ridge regression problem \cite{lukovsevivcius2012practical}. In this way, training does not require either backpropagation or gradient descent, which is usually problematic in time series forecasting \cite{bengio1994learning}.

Once the networks are trained, we integrate the Ca-ESN predictions in the formalism of controllers \eqref{eq:PD_control}-\eqref{eq:mpc}. In the PID controller, we use the maximum of the predicted observable for the uncontrolled system in the horizon $\tau_\mathrm{hor}$, as the control variable, $k(t) = \mathrm{max}(\hat{k}_{ESN}(t,t+\tau_\mathrm{hor}))$. Because we use the maximum, we simplify the controller by neglecting the derivative and integral terms ($K_d=K_i=0$). A schematic implementation is shown in Algorithm \ref{alg:PID-ESN}.
In MPC, the network provides the model: $\mathbf{f}_\mathrm{ESN}(\mathbf{q}(t_i), \mathbf{u}(t_i))$ predicts the future evolution of the controlled system in \eqref{eq:mpc}\footnote{The code is implemented in JAX \cite{jax2018github}, and publicly available on \href{https://github.com/MagriLab/Control-aware-ESN}{GitHub}.}. 

\begin{algorithm}
\caption{PID control with echo state networks}
\label{alg:PID-ESN}
\begin{algorithmic}


\For{$t \gets t_0$ to $t_N$} \Comment{Every control step}

$\mathbf{\hat{q}}(t) = \mathbf{q}(t)$

$k(t) = \mathbf{g}(\mathbf{\hat{q}}(t))$

\For{$t' \gets t$ to $t+\tau_\mathrm{hor}$} \Comment{Evaluate the ESN up to $\tau_\mathrm{hor}$}

$\mathbf{\hat{q}}(t'+dt) = \mathbf{f}_\mathrm{ESN}(\mathbf{\hat{q}}(t'), \mathbf{u}(t'))$

$k(t) = \max(k(t), \mathbf{g}(\mathbf{\hat{q}}(t'+dt)))$ \Comment{Save maximum of the observable}

\EndFor

$\mathbf{u}(t) = c(k(t))$

$\mathbf{q}(t+dt) = \mathbf{f}(\mathbf{q}(t), \mathbf{u}(t))$ 
\Comment{Apply control}


\EndFor
\end{algorithmic}
\end{algorithm}

\section{Results}

\begin{figure}[ht!]
    \centering
    \includegraphics[width=.60\textwidth]{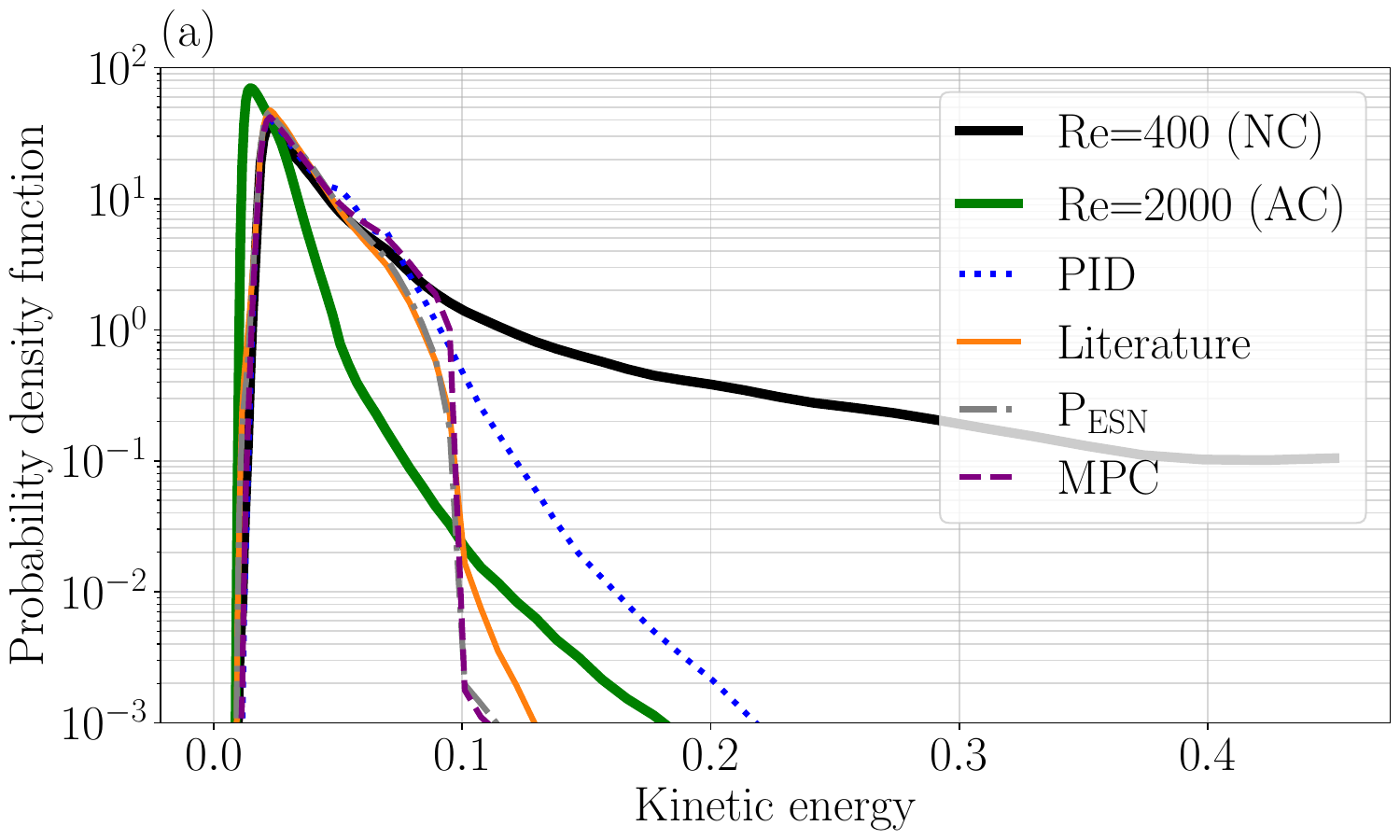}
    \includegraphics[width=1\textwidth]{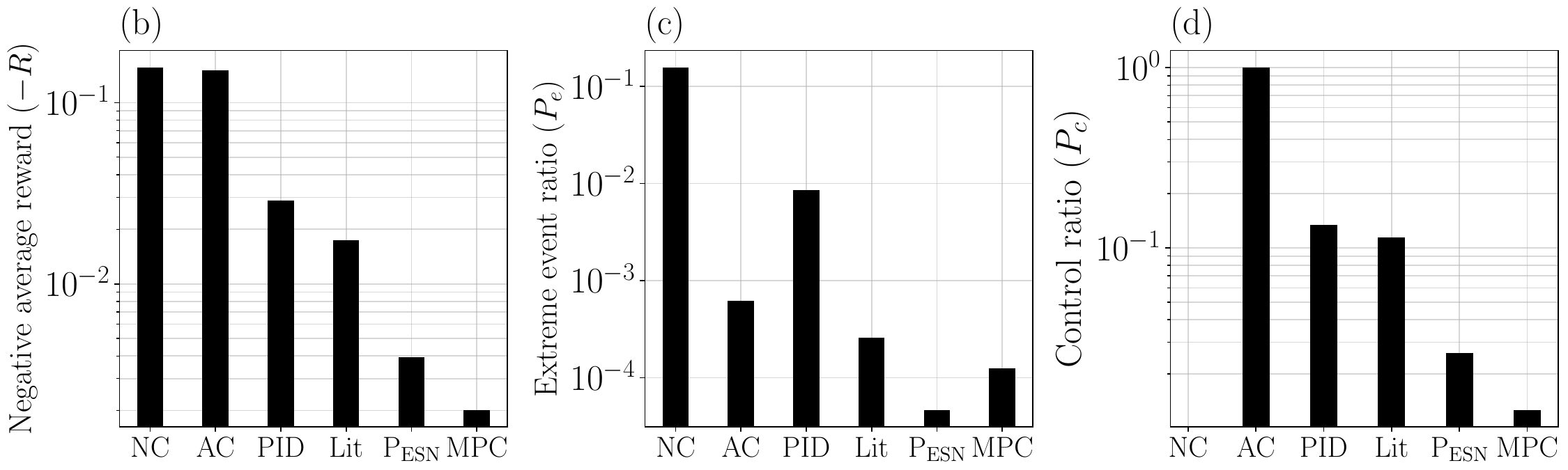}
\caption{(a) Probability density function of the kinetic energy, (b) average reward, (c) extreme event ratio and (d) control ratio for different control strategies.} 
    \label{fig:Results}
\end{figure}

To test the Ca-ESN, we consider the MFE, which is a qualitative model of turbulence \cite{moehlis2004low}. The dynamics are governed by the non-dimensional incompressible Navier-Stokes equations 
\begin{equation}
    \nabla\cdot\mathbf{v} = 0,\qquad \frac{d\mathbf{v}}{dt} + (\mathbf{v}\cdot\nabla)\mathbf{v} = - \nabla p + \frac{1}{\mathrm{Re}}\Delta \mathbf{v}  + \mathbf{f},
    \label{eq:NS}
\end{equation}
where $\mathbf{v}$ is the velocity, $p$ is the pressure, Re is the Reynolds number, and $\mathbf{f}$ is the body forcing that sustains turbulence. The MFE model is generated by  projecting \eqref{eq:NS} onto compositions of Fourier modes, which spawns nine nonlinear ordinary differential equations for the amplitudes of the modes, $q_i(t)$, which become the unknowns of the system \cite{moehlis2004low}. 
To integrate the equations, we use the same parameters and boundary conditions as \cite{racca2022data}.
For a wide range of Reynolds numbers, the system displays chaotic dynamics characterized by extreme events of the kinetic energy, $k(t)=\frac{1}{2}\sum_{i=1}^{9}q_i^2(t); \; k(t) > k_e =  0.1$ (Fig. \ref{fig:extreme_dyn}, \ref{fig:Results}a), whose probability
decreases with Re \cite{racca2022data}.
We analyse the highly extreme $Re=400$ regime, and suppress extreme events by temporarily increasing the Reynolds number, i.e, $\mathbf{u}(t_i)=Re(t_i)=2000$, following \cite{racca2022data}. We set the control penalisation term,  $r(t_{i,c}) = - 0.15$, for the two regimes to be equally desirable on average ($R_{400} \simeq R_{2000}$), so that a combination of the two is found by the controller.
In this control setting, (i) the PID controller is activated when $c(t)>k_c$, where $k_c$ is optimised instead of the proportional multiplier ($K_p=1$), and (ii) the optimisation of the discrete MPC problem is solved through complete search. 


We test the control strategies on 100,000 time series of length 20 Lyapunov times (LT)\footnote{The Lyapunov time is the inverse of the Lyapunov exponent of the system, which measures the average divergence of close-by trajectories in chaotic dynamics. In the MFE, 1LT $=0.0163^{-1}$ time units \cite{racca2022data}.}.
One representative time series, in which the event is suppressed through the Ca-ESN proportional controller, is shown in Fig. \ref{fig:extreme_dyn}b.
The network is trained on 50 time series only, and evolves for an horizon $\tau_\mathrm{hor}=4$ LTs, with a control horizon $N_\mathrm{opt}dt = 1$ LT, which are selected as a trade-off between the computational cost of the prediction and its capability of suppressing extreme events.
The controllers act on the system every 10 time units, which result in $\mathcal{O}(10^7)$ control steps analysed for each strategy. 

Figure \ref{fig:Results} shows the quantitative results. 
First, the standard PID controller decreases the number of extreme events by more than one order of magnitude compared to the uncontrolled (NC) system (Fig. \ref{fig:Results}c). 
Second, integrating the Ca-ESN in the proportional controller ($\mathrm{P}_\mathrm{ESN}$) and model predictive control (MPC), 
markedly improves the reward of the controllers with respect to the literature (Lit) \cite{racca2022data} (Fig. \ref{fig:Results}b). The Ca-ESN controllers decrease the occurrence of extreme events, while requiring significantly fewer actions than other methods (Fig. \ref{fig:Results}c,d). This shows that employing the networks is highly beneficial for the suppression of the events.  
Third, the Ca-ESN controllers decrease the occurrence of extreme events with respect to always controlling (AC) the system (Fig. \ref{fig:Results}c). This indicates that optimally selecting the active control strategy is more effective than passive control.

\section{Conclusions}

We propose the control-aware echo state network (Ca-ESN) to integrate ESNs into the formalism of conventional control strategies to suppress extreme events in chaos. 
The architecture is demonstrated on a turbulent flow, in which we combine the networks with PID controllers and model predictive control.
We show that the Ca-ESN (i) decreases the occurrence of extreme events up to two orders of magnitude with respect to both the uncontrolled and standard PID scenarios, (ii) requires an order of magnitude fewer actions to do so, 
and (iii) is more effective in suppressing the events than controlling the system at all times (passive control). 
This work opens up opportunities for the efficient control of extreme nonlinear dynamics from data, without the knowledge of the governing equations.

\begin{ack}
A. R. is supported by the Eric and Wendy Schmidt AI in Science Postdoctoral Fellowship, a Schmidt Futures program.  L. M. gratefully acknowledges financial support from the ERC Starting Grant PhyCo 949388 and from the UKRI AI for Net Zero grant EP/Y005619/1.
\end{ack}











\end{document}